\documentclass[conference]{IEEEtran}
\IEEEoverridecommandlockouts
\usepackage{cite}
\usepackage{times} % 使用 Times New Roman 字体
\usepackage[T1]{fontenc} % 确保正确编码
\usepackage{amsmath,amssymb,amsfonts}
\usepackage{algorithmic}
\usepackage{graphicx}
\usepackage{textcomp}
\usepackage{xcolor}
\usepackage{multirow} 
\usepackage{caption}     % 控制图和表的标题样式
\usepackage{subcaption}  % 处理子图
\def\BibTeX{{\rm B\kern-.05em{\sc i\kern-.025em b}\kern-.08em
T\kern-.1667em\lower.7ex\hbox{E}\kern-.125emX}}
\begin{document}

\title{Generative Image Coding with Diffusion Prior \\
\thanks{This work is partially sponsored by the National Key Research and Development Program of China (2024YFB4505704).}
}

% \author{Anonymous ICME submission}
\author{\IEEEauthorblockN{Jianhui Chang}
\IEEEauthorblockA{\textit{China Telecom Cloud Computing Research Institute} 
% } }
\\ \textit{changjh1@chinatelecom.cn} }}

\maketitle

\begin{abstract}
As generative technologies advance, visual content has evolved into a complex mix of natural and AI-generated images, driving the need for more efficient coding techniques that prioritize perceptual quality. Traditional codecs and learned methods struggle to maintain subjective quality at high compression ratios, while existing generative approaches face challenges in visual fidelity and generalization. To this end, we propose a novel generative coding framework leveraging diffusion priors to enhance compression performance at low bitrates. Our approach emplopys a pre-optimized encoder to generate generalized compressed-domain representations, integrated with pretrained model’s internal features via a lightweight adapter and an attentive fusion module. This framework effectively leverages existing pretrained diffusion models and enables efficient adaptation to different pretrained models for new requirements with minimal retraining costs. We also introduce a distribution renormalization method to further enhance reconstruction fidelity. Extensive experiments show that our method: (1) outperforms existing methods in visual fidelity across low bitrates, (2) improves compression performance by up to 79\% over H.266/VVC, and (3) offers an efficient solution for AI-generated content while being adaptable to broader content types.

\end{abstract}

\begin{IEEEkeywords}
Generative coding, image compression, latent diffusion models, AI-generated content
\end{IEEEkeywords}

\section{Introduction}
\label{sec:intro}

With the rapid advancement of generative technologies, visual content has evolved from predominantly natural images to a diverse blend of natural and generated images, driving the need for more efficient image coding techniques that optimize subjective perceptual quality. Transform-based traditional coding standards (\textit{e.g.}, HEVC~\cite{sullivan2012overview}, VVC~\cite{bross2021developments}) and end-to-end learned coding approaches~\cite{balle2020nonlinear,liu2023learned} tend to face challenges in maintaining subjective reconstruction quality at high compression ratios, primarily due to data degradation and insufficient utilization of external priors.

% 再想想怎么改这块的说法是，收窄一点范围）
Recent generative image coding approaches~\cite{mentzer2020high,chang2023semantic} leverage the powerful ability of generative models to capture complex data distributions, enabling high reconstruction quality at ultra-low bitrates. While promising, current techniques mainly based on GANs~\cite{goodfellow2014generative} are limited by their generative capacity, particularly in preserving realistic textures and structural details. Moreover, the generalizability of current generative coding methods~\cite{Chang2022conceptual,chang2023semantic} is limited, as compression performance is closely tied to the model's training scope. 
These methods typically require end-to-end rate-distortion (RD) optimization for specific scene data (\textit{e.g.,} AI-generated content or natural scenes), making training resource-intensive. Additionally, for different scenarios, retraining the entire model is often necessary, which incurs significant overhead and decoding challenges across devices.

% Models optimized for specific applications, such as AI-generated content or natural scenes, may produce compressed representations that are incompatible across different scenarios, leading to decoding challenges across devices.

% Recently, diffusion models have demonstrated remarkable capabilities in the image generation field, particularly in text-to-image tasks. The images generated by these models are typically rich in textures and well-structured, indicating that pretrained generative models contain substantial prior knowledge about visual content construction. This enables them to effectively capture and generate content from low-level textures to mid-level edges and high-level semantics. 

Diffusion models have recently achieved notable success in image generation, particularly in text-to-image tasks. Their ability to produce visually rich and well-structured images demonstrates substantial prior knowledge about visual content construction, spanning from low-level textures to high-level semantics. Existing works on diffusion-based compression follow different design paradigms. One line of research~\cite{hoogeboom2023high,ghouse2023residual} leverages diffusion models as post-processors atop existing codecs, focusing on enhancing decoded image quality rather than generative coding. Another line~\cite{lei2023text+sketch,careil2023towards} employs diffusion models as decoders conditioned on image-derived representations. 
% yang2024lossy,
% For instance, diffusion-based decoders conditioned on compressed latents~\cite{yang2024lossy} require relatively high bitrates (\textit{e.g.}, $>$ 0.3 bpp) to achieve competitive perceptual fidelity. 
For instance, text-conditioned diffusion decoders~\cite{lei2023text+sketch,careil2023towards} have been explored for exceptionally low bitrates ($<$ 0.01 bpp), excelling at preserving high-level semantic consistency, but often sacrificing visual fidelity. Additionally, the concurrent work PerCo~\cite{careil2023towards} utilizes vector-quantized codebooks to represent images alongside textual information. However, such methods typically require separate training of the auxiliary encoder with codebooks of different sizes, along with finetuning of the diffusion model for varying target rates and content, limiting their flexibility and efficiency in adapting to diverse compression needs.

To this end, this paper proposes a generative coding framework that leverages powerful diffusion priors to enhance compression performance, focusing primarily on improving perceptual fidelity across a broad range of low bitrates, typically around 0.01 to 0.2 bpp. We leverage pretrained latent diffusion models~\cite{rombach2022high} as generative decoders, while optimizing the encoder via an pretext end-to-end learned compression task, independent of specific generative models. A lightweight adapter is introduced to ensure effective domain adaptation, providing a flexible and compatible solution across different models. To further improve reconstruction fidelity and compression efficiency, we incorporate a cross-attention mechanism for better alignment between compressed latents and internal features, and a distribution renormalization method to mitigate reconstruction distortion.
The primary contributions of this work are as follows:
\begin{itemize}
    \item We propose a generative coding framework leveraging powerful diffusion priors to achieve efficient human-centered compression. The framework includes a lightweight adapter and an attentive fusion module that enable effective domain adaptation and ensure compatibility across various pretrained latent diffusion models.
    \item We introduce a cross-attention mechanism to refine the integration of compressed latents and the pretrained model's internal features, along with a distribution renormalization method to enhance reconstruction fidelity, both of which further boost compression performance.
    \item Extensive experiments show that our method achieves up to a 79\% improvement in compression performance compared to VVC, while demonstrating versatility across natural and AI-Generated scenarios with different pretrained diffusion models.
\end{itemize}

\begin{figure*}[t]  % htbp specifies the preferred placement (here, top, bottom, page)
\vspace{-20pt}
    \centering
    \includegraphics[width=0.88\textwidth]{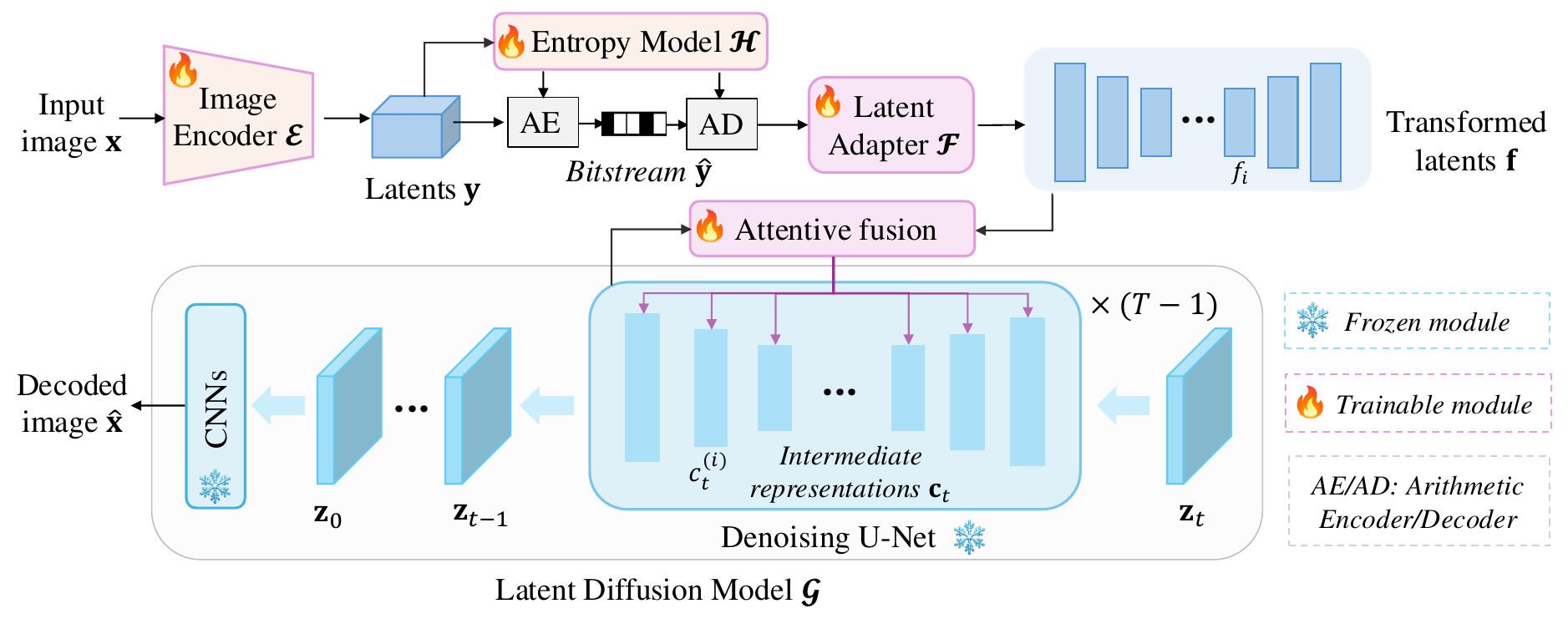} % Change 'your_image_file.png' to the actual file
    \vspace{-2mm}
    \caption{Overview of the proposed generative image coding framework.}
    \label{Fig:frame}  % Label for referencing in the text
\vspace{-14pt}
\end{figure*}

\section{Proposed Method}
The main framework of the proposed generative image coding method is shown in Fig.~\ref{Fig:frame}. 
It encompasses several key components: (1) an image encoder $\mathcal{E}$ and an entropy model $\mathcal{H}$, which collaborate to encode images into rate-constrained latents; (2) a latent adapter $\mathcal{F}$ equipped with an attentive fusion module, tasked with transforming the compressed latents into control signals; and (3) a latent diffusion model~\cite{rombach2022high} (LDM) $\mathcal{G}$, which serves as a robust prior for realistic image reconstruction.
% % 
% It comprises these components: the image encoder $\mathcal{E}$ and entropy model $\mathcal{H}$ for encoding images to rate-constraint latents, the latent adapter $\mathcal{F}$ with attentive fusion module for transforming compressed latents to control signals for subsequent generative models, and the latent diffusion model~\cite{rombach2022high} $\mathcal{G}$ as a robust prior, enabling realistic image reconstruction.
Our objective is to harness the capabilities of the pretrained model $\mathcal{G}$ for image reconstruction by integrating compressed latents with its internal priors through a lightweight $\mathcal{F}$ and attentive fusion, thus enabling accurate guidance of the reconstruction process by compressed signals.
% This allows the pretrained latent diffusion model to be controlled by compressed signals, guiding the image reconstruction process.

At the encoder side, to ensure flexibility and generalization, the pre-optimized encoder $\mathcal{E}$ is used to transform the input image $\mathbf{x}$ into a unified compressed-domain latent representation $\mathbf{y}$, defined as $\mathbf{y} = \mathcal{E}(\mathbf{x})$.
After transformation, the latents $\mathbf{y}$ are quantized and entropy-encoded into bitstreams. On the decoder side, the decoded latents $\hat{\mathbf{y}}$ are transformed by the latent adapter $\mathcal{F}$ into a feature set $\mathbf{f}$. This set is then aligned and fused with the intermediate representations $\mathbf{c}_t$ of the denoising U-Net at each time step $t$ within the latent diffusion model. This integration of encoded information steers the generative process, enabling the synthesis of a high-fidelity reconstructed image $\hat{\mathbf{x}}$.
% After transformation, the latents $\mathbf{y}$ are quantized and entropy-encoded into bitstreams. On the decoder side, the decoded latents $\hat{\mathbf{y}}$ are transformed by the latent adapter $\mathcal{F}$ to a feature set $\mathbf{f}$, which is aligned and fused with the pretrained model's internal features $\mathbf{c}$, incorporating the encoded information to control the generative process and synthesize a high-fidelity reconstructed image $\hat{\mathbf{x}}$.
% 
In the latent diffusion model, the generative process begins with random noise $\mathbf{z}$ sampled from a prior Gaussian distribution $\mathcal{N}(0, 1)$ and is conditioned on the compressed information $\hat{\mathbf{y}}$, expressed as:
\begin{equation}
    \hat{\mathbf{x}}=\mathcal{G}(\mathbf{z},\hat{\mathbf{y}}),   \hat{\mathbf{x}}\sim P(\hat{\mathbf{x}} \mid \hat{\mathbf{y}}).
\end{equation}
Since the compressed information $\hat{\mathbf{y}}$ provides strong spatial guidance, the target distribution $P(\hat{\mathbf{x}})$ is primarily determined by $\hat{\mathbf{y}}$ and the internal diffusion priors, allowing us to achieve high-fidelity reconstruction results for the compression task.
% 是否在这里就解释一下DDIM？
% 
\subsection{Latent Adapter and Attentive Fusion}
% Adapters have been widely used in computer vision for incremental learning and domain adaptation~\cite{de2021continual, wang2018deep}.
Various approaches have been explored to adapt pretrained models to different tasks or domains in computer vision, though they may not be ideally suited for image compression.
For instance, ControlNet~\cite{zhang2023adding} employs a trainable copy of a large-scale pretrained backbone to generate additional feature maps, incurring significant computational overhead. Another representative approach, T2I-Adapter~\cite{mou2024t2iadapter}, controls the color and structure of generated images but does not provide the required granularity to achieve the high perceptual reconstruction fidelity targeted by our method.
% required by generative image compression tasks.

% Adapters have been used in computer vision for incremental learning and domain adaptation~\cite{de2021continual,wang2018deep}. 
% However, existing approaches face limitations in training complexity and fine-grained control. 
% For instance, ControlNet\cite{zhang2023adding} employs a trainable copy of the large-scale pretrained backbone to learn additional feature maps, thereby introducing significant computational overhead. Mou \textit{et al.}~\cite{mou2024t2iadapter} introduce T2I-Adapter to control the color and structure of generated images, but they could not satisfy the perceptual fidelity requirements that generative image compression tasks pursue/making them less suitable for tasks requiring high fidelity and determinism.

To address these challenges, this paper designs a latent adapter and an attentive latent fusion module for high visual fidelity generative coding. The adapter aligns the compressed-domain latents $\hat{\mathbf{y}}$ with features from both the downsampling and upsampling modules of the U-Net, defined as $\mathbf{f} = \mathcal{F}(\hat{\mathbf{y}})$. Each feature extraction module includes a convolutional layer and two residual modules, producing features $f_i$ aligned with the corresponding U-Net features $c_t^{(i)}$ at each time step $t$. 

\begin{figure}[t]  % htbp specifies the preferred placement (here, top, bottom, page)
\vspace{-5pt}
    \centering
    \includegraphics[width=0.99\linewidth]{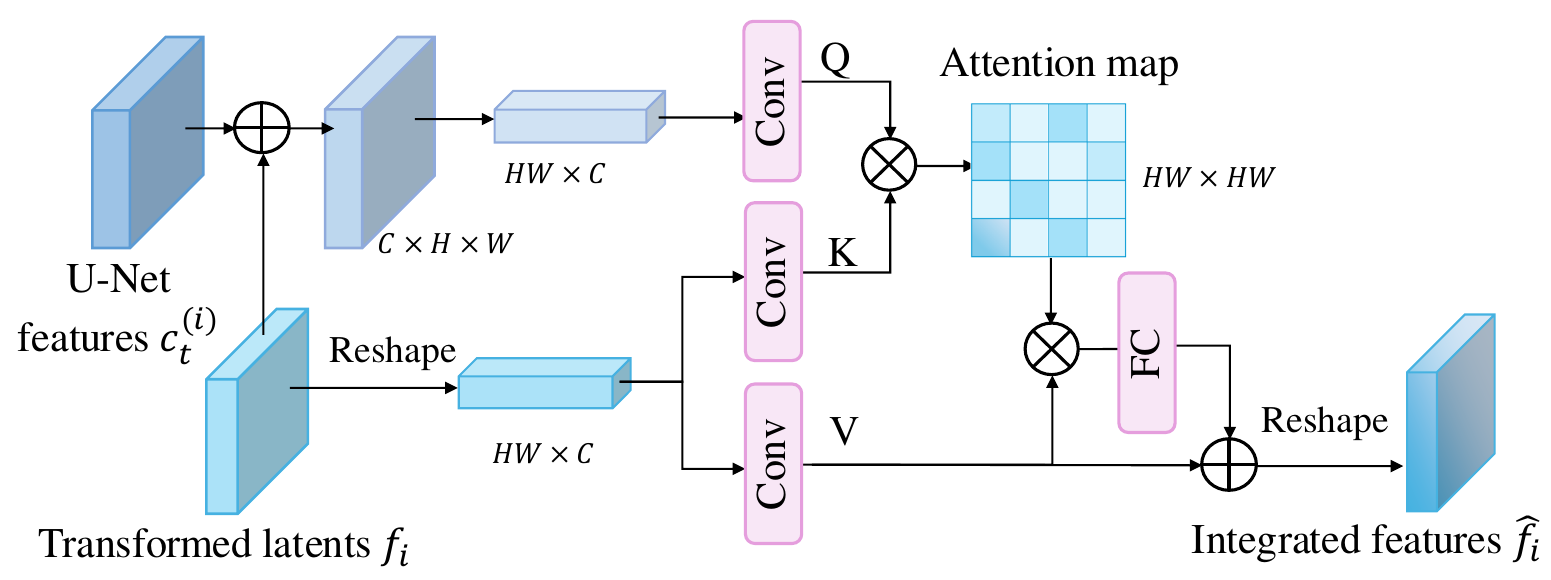} % Change 'your_image_file.png' to the actual file
    \vspace{-4mm}
    \caption{Attentive latent fusion based on spatial cross-attention.}
    \label{Fig:fusion}  % Label for referencing in the text
\vspace{-15pt}
\end{figure}

The transformed latents $\mathbf{f}$ and internal features $\mathbf{c}_t$ are then fused and integrated into the generative diffusion process for image reconstruction.
The commonly used additive fusion~\cite{mou2024t2iadapter,zhang2023adding} method assumes spatial alignment and overlooks contextual information, limiting reconstruction capability. Thus, we introduce an attentive fusion module to enhance fusion accuracy for improving image reconstruction fidelity based on spatial cross-attention. 
As shown in Fig.~\ref{Fig:fusion}, the U-Net intermediate features $c_t^{(i)}$ and transformed latents $f_i$ are summed to form the base feature $\hat{c}_t^{(i)} =  c_t^{(i)} + f_i$, enhancing spatial correlation. The transformed latents $f_i$, derived from compressed information, act as context vectors. By reshaping their dimensionality to $HW\times C$, we achieve a linearization of the spatial dimensions, allowing each feature vector to contribute independently to the attention mechanism. Three $1\times1$ convolution layers then map $\hat{c}_t^{(i)}$ to $\mathbf{Q}$, and reshape $f_i$ to $\mathbf{K}$ and $\mathbf{V}$. With $FC$ denoting a fully connected layer, the fused feature $\hat{\mathbf{f}}_i$ is computed as:
\begin{equation}
\label{eq_chap5:atten_v}
    \hat{\mathbf{f}}_i = \mathbf{V} + FC(Attn(\mathbf{Q}, \mathbf{K}, \mathbf{V}) ),
\end{equation}
\begin{equation}
\label{eq_chap5:atten_weight}
    {Attn}(\mathbf{Q}, \mathbf{K}, \mathbf{V}) = {Softmax}\left(\frac{\mathbf{QK}^{\top}}{\sqrt{C}}\right) \cdot \mathbf{V}.
\end{equation}
The cross-attention mechanism facilitates fusion by capturing correlations between compressed information and the generative model's internal features at each spatial location. 
% This approach retains the inherent information and integrates contextual information from other locations. 
Consequently, it enables the latent diffusion model to obtain more precise spatial details of the transformed latents derived from target images, effectively improving the accuracy of reconstructed images.
% This preserves inherent feature information, integrates contextual compressed information, and enhances spatial relationships, improving image reconstruction accuracy.

\subsection{Optimization Strategy}
Pretrained generative models usually provide strong priors for visual reconstruction, but retraining them is costly and risks degrading these priors. To avoid this, we keep the pretrained latent diffusion model $\mathcal{G}$ frozen and employ a two-stage optimization strategy: first optimizing the encoder $\mathcal{E}$ and entropy estimation model $\mathcal{H}$ through a pretext task, then freezing them and fine-tuning the lightweight adapter $\mathcal{F}$ and the attentive fusion module.

(1) \textbf{Encoder Optimization.} During the first stage, a learned image compression task~\cite{cheng2020learned} is employed for pretext optimization. This involves an encoder $\mathcal{E}$, an auxiliary decoder, and an entropy model $\mathcal{H}$. This auxiliary decoder is introduced specifically for the pretext optimization and will be discarded after the optimization is completed.
To achieve variable-rate coding with a unified encoder and entropy model, we utilize channel-wise scaling and quantization techniques, as detailed in \cite{cui2021asymmetric}.
For rate-distortion optimization~\cite{shannon1948mathematical,balle2019end}, each Lagrange multiplier $\lambda_s$ is associated with a set of quantization parameters for rate level $s$. The rate constraint $\mathcal{L}^{s}_{rate}$ is determined by $\mathcal{H}$. Using MS-SSIM \cite{wang2003multiscale} to measure the reconstruction distortion $\mathcal{L}^{s}_{dist}$, the pretext optimization objective is formulated as follows:
\begin{equation}
\label{eq_chap5:loss_rds}
    \mathcal{L}_{pretext} = \sum ^{L-1}_{s=0} \mathcal{L}^{s}_{rate}+\lambda_s \mathcal{L}^s_{dist}.
\end{equation}
During training with Eq.~(\ref{eq_chap5:loss_rds}), the auxiliary decoder is updated alongside $\mathcal{E}$ and $\mathcal{H}$ via stochastic gradient descent. Upon completion of the pretext optimization, only the parameters of $\mathcal{E}$ and $\mathcal{H}$ are retained and remain fixed for the subsequent training phase with other modules.
% For subsequent training and inference, only the encoder and entropy model are retained, with the decoder $\mathcal{S}$ discarded. 

(2) \textbf{Adapter Optimization.} After the first-stage training, the second stage targets the optimization of the latent adapter $\mathcal{F}$ and the attentive fusion module with $\mathcal{E}$, $\mathcal{H}$ and $\mathcal{G}$ kept fixed.
% 
% In the diffusion process, the latents $\mathbf{z}_0$ are derived from the outer encoder of the pretrained LDM for an input image $\mathbf{x}$. Noise is added to $\mathbf{z}_0$ over timesteps $t$, generating the noisy latents $\mathbf{z}_t$. 
In the diffusion process, latents $\mathbf{z}_0$ from the input image $\mathbf{x}$ are perturbed with noise at each timestep $t$ to produce $\mathbf{z}_t$.
Image reconstruction is performed iteratively in the reverse diffusion process by denoising $\mathbf{z}_t$ using a U-Net noise estimator $\epsilon_\theta$, conditioned on the transformed compressed feature set $\mathbf{f}$. 
% 
% Image generation is achieved by iteratively denoising with a U-Net noise estimator $\epsilon_\theta$ with the condition of transformed compressed feature set $\mathbf{f}$. 
The denoising loss function is defined as follows:
\begin{equation}
\label{eq_chap5:loss_diffusion}
\mathcal{L}_{adp}=\mathbb{E}_{\mathbf{z}_0, t, \mathbf{f}, \epsilon \sim \mathcal{N}(0,1)}\left[\left\|\epsilon-\epsilon_\theta\left(\mathbf{z}_t, t, \mathbf{f}\right)\right\|_2^2\right].
\end{equation}

In the training phase, input images are encoded by $\mathcal{E}$, transmitted, and transformed by $\mathcal{F}$ to $\mathbf{f}$, which is then integrated into the denoising process via the fusion module. With other modules held fixed, $\mathcal{F}$ and the fusion module undergo fine-tuning, enabling the utilization of the pre-trained model $\mathcal{G}$ for the image compression task upon completion.
\subsection{Fidelity Enhancement}
The proposed generative coding method successfully reconstructs images while maintaining consistent visual semantics and structure. However, distortions in color distribution highlight the need for improved fidelity. In image processing, color moments—mean ($\mu$) and standard deviation ($\sigma$)—are fundamental metrics for characterizing color distribution, with the mean capturing brightness and the standard deviation reflecting contrast through pixel value variation. Style transfer research~\cite{huang2017arbitrary} demonstrates that tuning instance normalization in DNN feature spaces retains content while transferring color styles.
% Additionally, research in style transfer~\cite{huang2017arbitrary} has shown that adjusting instance normalization statistics within deep neural network feature spaces can preserve content structure while incorporating color styles. 
Extending this insight, we propose a method to correct color deviations in the synthesized images' \textit{pixel domain} by aligning their channel-wise mean and standard deviation with the original image, thereby enhancing reconstruction fidelity.

% The proposed generative coding method effectively reconstructs images with consistent visual semantics and structure. However, color distribution distortions suggest the need for enhanced fidelity. In the field of image processing, color moments—mean ($\mu$) and standard deviation ($\sigma$)—are key metrics for describing color distribution, where the mean reflects brightness and the standard deviation represents contrast through pixel value variation. Additionally, studies~\cite{huang2017arbitrary} in style transfer have demonstrated that adjusting instance normalization statistics within deep neural network feature spaces enables the preservation of content structure while incorporating color styles. Building on this observation, this paper introduces a method that corrects color deviations in synthesis images by adjusting channel-wise statistics—mean ($\mu$) and standard deviation ($\sigma$)—to match the original image's distribution, improving reconstruction fidelity.
% 
% To this end, we introduce a method that corrects color deviations by adjusting channel-wise statistics—mean ($\mu$) and standard deviation ($\sigma$)—to match the original image's distribution inspired by style transfer~\cite{huang2017arbitrary}. 
% 
\begin{figure*}[t]  % htbp specifies the preferred placement (here, top, bottom, page)
\vspace{-15pt}
    \centering
    \includegraphics[width=0.89\textwidth]{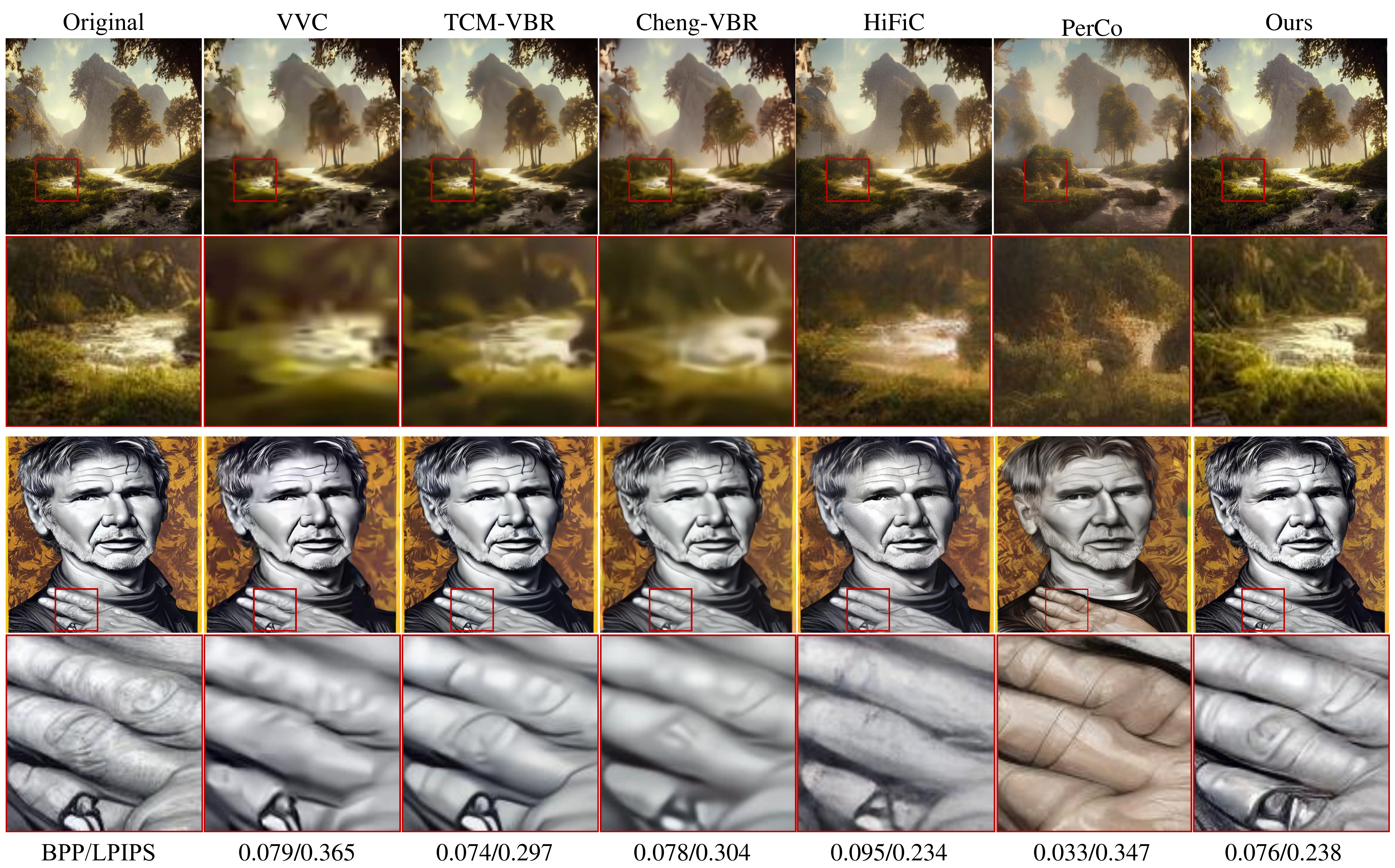} % Change 'your_image_file.png' to the actual file
    % \vspace{-2mm}
    \caption{Qualitative comparisons of VVC, Cheng-VBR, TCM-VBR, HiFiC, PerCo and ours on the AIGC dataset.}
    \label{Fig:sub_aigc}  % Label for referencing in the text
    \vspace{-10pt}
\end{figure*}
Specifically, to minimize reconstruction distortion, the color distribution parameters of the reconstructed image should align with those of the original. The mean ($\mu$) and standard deviation ($\sigma$) are computed for each channel, with $\mu = {\mu_R, \mu_G, \mu_B}$ and $\sigma = {\sigma_R, \sigma_G, \sigma_B}$. For efficient transmission, these parameters are quantized to $\hat{\mu}$ and $\hat{\sigma}$ with a step size $\Delta$.
On the decoder side, the initial reconstruction $\hat{\mathbf{x}}$ is renormalized for fidelity enhancement using the transmitted parameters $\hat{\mu}$ and $\hat{\sigma}$ as follows:
\begin{equation}
    \hat{\mathbf{x}}_{norm} = (\frac{\hat{\mathbf{x}}-\hat{\mu}}{\hat{\sigma}})\sigma + \mu,
\end{equation}
where $\hat{\mathbf{x}}_{norm}$ represents the final decoded image after improving color fidelity.
The proposed approach significantly improves color accuracy in reconstructed images. Moreover, we find that using statistics from smaller regions leads to more accurate color correction. This approach is further extended to block-level correction, where statistical parameters are transmitted and applied for each image block, ensuring more precise color alignment with negligible bitrate cost.

\section{Experiments}
\subsection{Experimental Settings}
\paragraph{Datasets} Artificial Intelligence Generated Content (AIGC) has recently become a key AI research focus. The statistics of AIGC data align well with generative model priors, making it well-suited for generative coding. Thus, this study utilizes the DiffusionDB dataset \cite{wang2022diffusiondb}, containing 14 million high-quality, predominantly anime-style images generated from real user prompts. For experiments, 50,634 images are randomly selected for training and 200 for testing at a resolution of $512 \times 512$. 
Besides, to evaluate the method's compatibility with other pretrained diffusion models on natural scenes, 100,000 images of OpenImages dataset\footnote{https://storage.googleapis.com/openimages/web/index.html} are used to fine-tune the latent adapter. The Kodak dataset\footnote{https://r0k.us/graphics/kodak/} is used to assess compression performance on real-world images. 

\paragraph{Implementation Details}
All experiments are conducted using PyTorch on four NVIDIA Tesla 32G-V100 GPUs. The Adam optimizer is used with a learning rate of $1e-5$. The first-stage model follows Cheng et al.~\cite{cheng2020learned} with 128 latent channels, is trained for 600 epochs with Lagrange multipliers \{50.0, 16.0, 3.0, 1.0, 0.5, 0.25, 0.1, 0.05, 0.01, 0.005\} and a batch size of 96, starting from pretrained CompressAI\footnote{https://interdigitalinc.github.io/CompressAI/} parameters. The pretrained latent diffusion model is sourced from Stable Diffusion v1.4\footnote{https://huggingface.co/CompVis/stable-diffusion-v1-4} with an input of $512 \times 512$ and latent dimensions of $4 \times 64 \times 64$. 
The latent adapter network and fusion module, shared across bitrates, are fine-tuned for 10 epochs with a batch size of 8. During inference, images are generated using the DDIM deterministic sampling schedule~\cite{song2020denoising} with 10 sampling steps. 
The test results are obtained using a global random seed of 42, ensuring deterministic initialization of latent noise $\mathbf{z}$ for the denoising process. Varying the random seed produces a set of reconstructions that may reflect the uncertainty inherent in specific compressed latents~\cite{careil2023towards}. The standard deviation of LPIPS values across different random initializations is empirically below 0.01 in our results, indicating its impact on reconstruction results is negligible.
% 
% The test results are obtained using a global random seed of 42, ensuring deterministic latent noise $\mathbf{z}$ initialization for the denoising process. Varying the random seed allows us to obtain a set of reconstructions that reflect the uncertainty under specific compressed latents. The standard deviation of LPIPS values across different random initializations is generally below 0.01, which has negligible impact on the reconstruction results.
% suggesting that the reconstructions are visually stable with imperceptible differences. 
% (illustrated in Appendix xx)
% 
For fidelity enhancement, a block size of $64 \times 64$ is used, with parameters quantized to 6 bits. Encoding these parameters adds an average of $0.01$ bpp.
% The latent adapter network and fusion module, shared across bitrates, are finetuned for 10 epochs with a batch size of 8. At inference time, images are generated with the DDIM deterministic sampling schedule~\cite{song2020denoising}. We use 10 sampling steps to achieve good performance. The test results are obtained with global random seed 42.  The latent noise $\mathbf{z}$ sampled as the starting point for the denoising process will be deterministic with a fixed random seed. We also evaluate that different random seed will result in imperceptible difference measured  LPIPS and MS-SSMI < 0.01.(average standard variation)

\begin{figure}[t]
% \vspace{-15pt}
    \centering
    % 第一子图
    \begin{subfigure}[b]{0.482\linewidth}
        \centering
        \includegraphics[width=\linewidth]{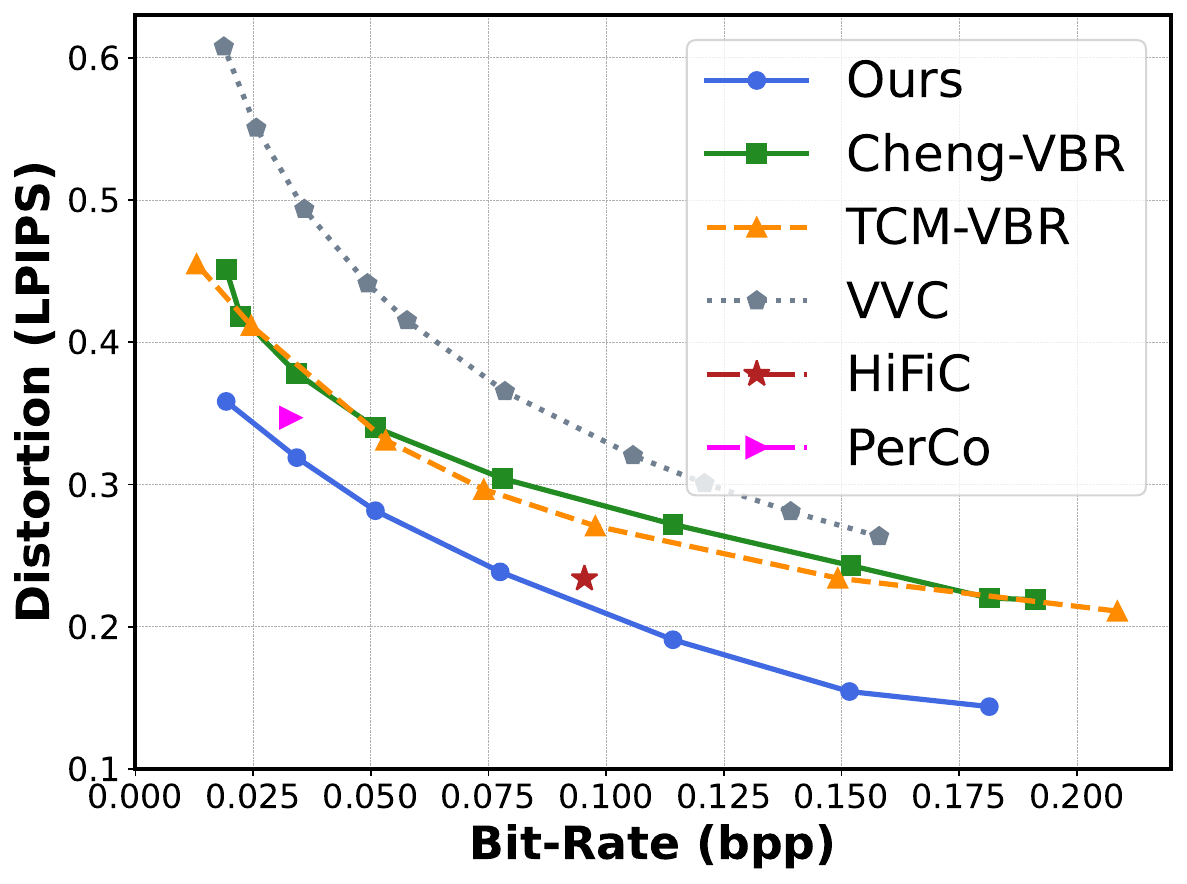}
        \caption{R-D results on DiffusionDB.}
        \label{subfig:comparison1}
    \end{subfigure}
    \hfill
    % 第二子图
    \begin{subfigure}[b]{0.49\linewidth}
        \centering
        \includegraphics[width=\linewidth]{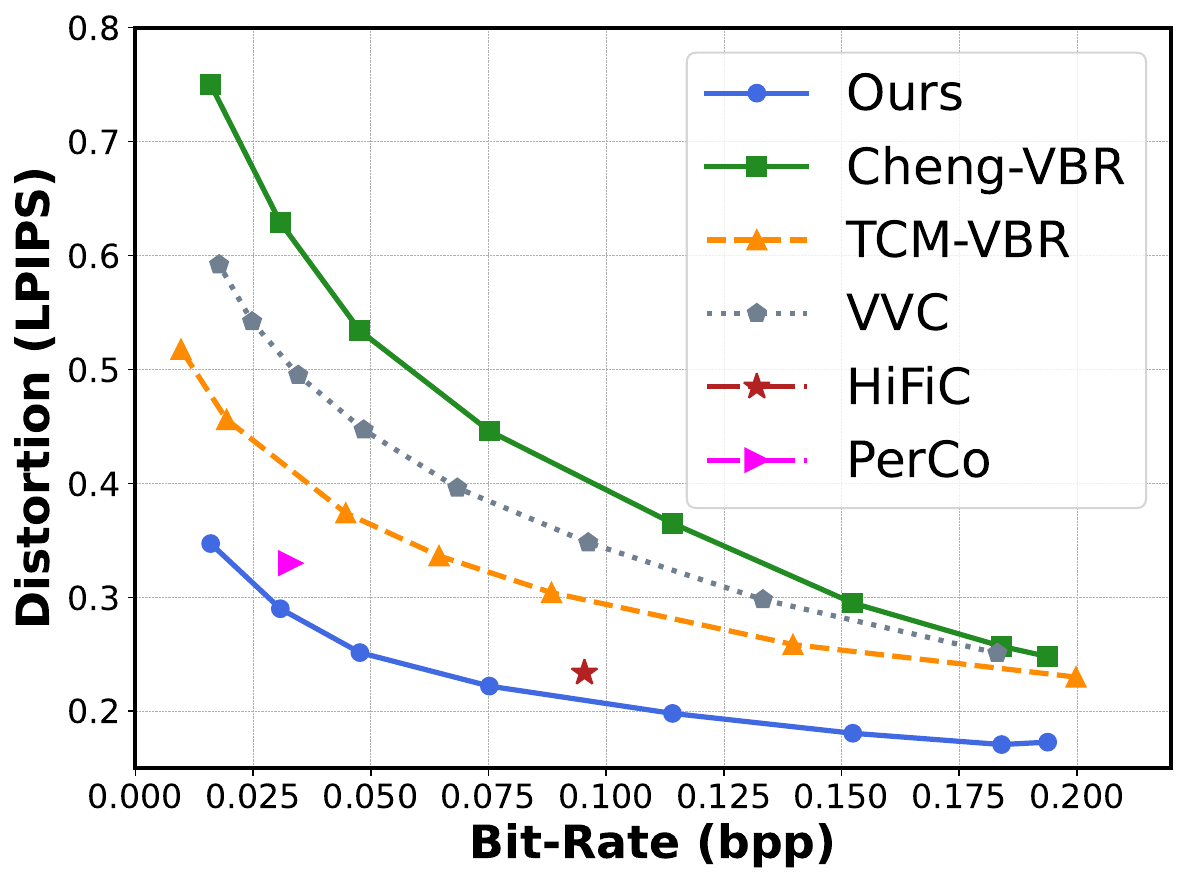}
        \caption{R-D results on Kodak.}
        \label{subfig:comparison2}
    \end{subfigure}
    \caption{The R-D comparisons on Kodak and the AIGC dataset, DiffusionDB. Lower LPIPS scores indicate higher fidelity.}
    \label{Fig:RD_all}
    \vspace{-20pt}
\end{figure}

\paragraph{Evaluation Metrics} Traditional coding methods typically focus on preserving signal fidelity, using metrics like PSNR and SSIM to evaluate pixel-level distortions. However, these metrics often struggle to reflect the perceptual quality. 
Likewise, dataset-level measures like FID and KID assess distribution discrepancies but are insufficient for evaluating image-level perceptual fidelity.
Therefore, we adopt the Learned Perceptual Image Patch Similarity (LPIPS) \cite{zhang2018unreasonable}, which measures feature-domain distortion between paired images, better aligns with human perception, and is widely recognized in the field~\cite{mentzer2020high,chang2023semantic}. Additionally, bits per pixel (bpp) is employed to evaluate rate performance. 
% Additional quantitative comparisons on PSNR and MS-SSIM are provided in Sec.~I of the supplementary.
% Given our focus on perceptual quality and fideility
% Some studies measure dataset-level distribution discrepancies, such as FID and KID. While useful, these metrics may not accurately reflect image-level perceptual fidelity. 
% \paragraph{Evaluation metrics.} Traditional coding methods usually preserve signal fidelity and employ traditional objective quality assessment methods to calculate pixel-level distortions, such as PSNR and SSIM. These methods may not be suitable for evaluating perceptual quality which generative coding methods usually target. Some work use distribution similarity metrics to evaluate visual quality, such as FID and KID, which are devised to compute distribution distortion at dataset-level and thus may not indicate image-level perceptual fidelty. Thus, we utilize the learned perceptual image patch similarity (LPIPS)\cite{zhang2018unreasonable} to evaluate image-level perceptual fidelity, which is also proved to better align with the way humans perceive images and is widely used. Besides, bits per pixel (bpp) is used to indicate the rate performance.

\subsection{Compression Performance Comparison}
\paragraph{Compared Methods}
The proposed method is compared with several representative baselines: (1) \textbf{VVC}, the Versatile Video Coding standard~\cite{bross2021developments}, using VTM-11.0 with intra-coding under common test conditions; (2) \textbf{Cheng-VBR}, a deep learning-based end-to-end compression method \cite{cheng2020learned}, used for pretext optimization in this study; (3) \textbf{TCM-VBR}, one of the state-of-the-art methods~\cite{liu2023learned} utilizing a learnable quantization scale matrix and training strategy similar to this work for variable bitrate coding; and (4) \textbf{HiFiC}, a generative compression method \cite{mentzer2020high}, trained at a fixed low bitrate. (5) \textbf{PerCo\footnote{https://github.com/Nikolai10/PerCo}}~\cite{careil2023towards}, one of the latest diffusion-based generative coding methods. The released checkpoints use Stable Diffusion v2.1, which is more advanced than ours.
% All learned methods are finetuned on the DiffusionDB dataset.

\begin{figure}[t]  % htbp specifies the preferred placement (here, top, bottom, page)
    \vspace{-15pt}
    \centering
    \includegraphics[width=0.99\linewidth]{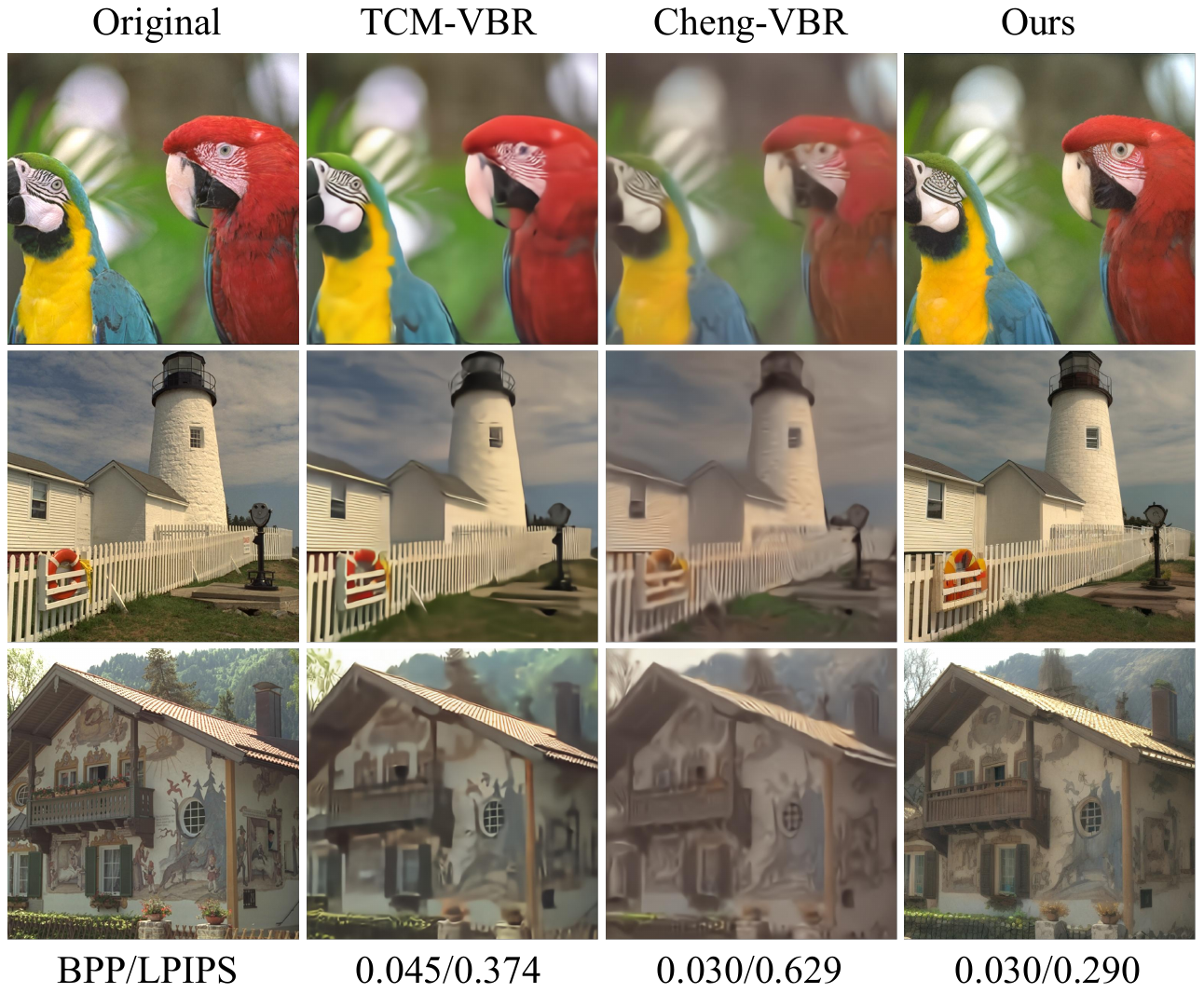} % Change 'your_image_file.png' to the actual file
    % \vspace{-1mm}
    \caption{Qualitative comparisons on the Kodak dataset.}
    \label{Fig:sub_kodak}  % Label for referencing in the text
    \vspace{-15pt}
\end{figure}
\paragraph{Quantitative Evaluation}
Fig.~\ref{Fig:RD_all} shows that the proposed generative coding method significantly outperforms VVC, Cheng-VBR \cite{cheng2020learned}, HiFiC, PerCo and TCM-VBR \cite{liu2023learned} in RD performance using LPIPS as the metric. Tab.~\ref{tab_chap5:BD-Rate} further quantifies this improvement with the Bjontegaard metric~\cite{bjontegaard2001calculation}, showing bitrate reductions of 67.8\% compared to VVC, 50.1\% compared to Cheng-VBR, and 40.1\% compared to TCM-VBR on the AIGC dataset. In terms of BD-LPIPS, the proposed method enhances reconstruction quality by 27.0\% over Cheng-VBR, 24.5\% over TCM-VBR, and 37.8\% over VVC.
% showing bitrate savings of 67.8\% over VVC, 50.1\% over Cheng-VBR, and 40.1\% over TCM-VBR on AIGC dataset. In terms of BD-LPIPS, the proposed method enhances reconstruction quality by 27.0\% over Cheng-VBR, 24.5\% over TCM-VBR, and 37.8\% over VVC. 
% Fig.~\ref{Fig:RD_all} shows that the proposed generative coding method significantly outperforms VVC, Cheng-VBR \cite{cheng2020learned}, HiFiC and TCM-VBR \cite{liu2023learned} in RD performance using LPIPS as the metric. Tab.~\ref{tab_chap5:BD-Rate} further quantifies this improvement with the Bjontegaard metric~\cite{bjontegaard2001calculation}, showing bitrate savings of 67.8\% to 79.2\% over VVC, 50.1\% to 81.6\% over Cheng-VBR, and 40.1\% to 69.9\% over TCM-VBR on AIGC and Kodak dataset. In terms of BD-LPIPS, the proposed method enhances reconstruction quality by 27.0\% to 45.8\% over Cheng-VBR, 24.5\% to 29.3\% over TCM-VBR, and 37.8\% to 39.4\% over VVC. 
TCM-VBR combines CNNs and transformers in its transform and entropy models to enhance both local and global feature capture, achieving better RD performance than Cheng-VBR and VVC. HiFiC's performance is constrained by the suboptimal generative capabilities of GANs, while PerCo introduces textual information that does not contribute to improving visual fidelity. In contrast, the proposed method leverages powerful diffusion priors and an attentive fusion mechanism, enabling more perceptually faithful reconstructions from compact latent representations. 
% Detailed ablation studies of the proposed methods are presented in Sec.~II of the supplementary. 
These results highlight the method's strong RD performance across a wide compression ratio range (100$\times$ to 2000$\times$), demonstrating its effectiveness. 
% TCM-VBR incorporates CNNs and transformers in its transform and entropy models to enhance both local and global feature capture, resulting in better RD performance than Cheng-VBR and VVC. The performance of HiFiC is limited by sub-optimial generative capability of GANs and PerCo introduces textual information which does not benefit visual fidelty improvement. In contrast, the proposed method leverages powerful diffusion prior and attentive fusion method, enabling more perceptually faithful reconstructions from compact latent representations.
% These results validate the method's strong RD performance across a wide compression ratio range (100$\times$ to 2000$\times$), demonstrating its effectiveness.
% 

\begin{table}[t]
\centering
\vspace{-10pt}
\caption{BD-Rate and BD-metric results relative to VVC, Cheng-VBR and TCM-VBR respectively. LPIPS is used as the distortion metric. }
% \vspace{-5pt}
\label{tab_chap5:BD-Rate}
\resizebox{0.99\linewidth}{!}{
\begin{tabular}{c|c|c|c|c}
\hline
Dataset                & Metric   & VVC      & Cheng-VBR & TCM-VBR  \\ \hline
\multirow{2}{*}{AIGC}  & BD-Rate  & -67.75\% & -50.14\%  & -40.08\% \\ \cline{2-5} 
                       & BD-LPIPS & -37.76\% & -27.03\%  & -24.46\% \\ \hline
\multirow{2}{*}{Kodak} & BD-Rate  & -79.24\% & -81.62\%  & -69.93\% \\ \cline{2-5} 
                       & BD-LPIPS & -39.42\% & -45.80\%  & -29.34\% \\ \hline
\end{tabular}}
\vspace{-12pt}
\end{table}

\paragraph{Qualitative Evaluation}
Fig.~\ref{Fig:sub_aigc} presents the subjective reconstruction results of the proposed generative compression method alongside those of VVC, Cheng-VBR~\cite{cheng2020learned}, TCM-VBR~\cite{liu2023learned}, HiFiC~\cite{mentzer2020high} and PerCo~\cite{careil2023towards}. VVC exhibits noticeable block artifacts, while TCM-VBR and Cheng-VBR suffer from excessive smoothing and blurring. HiFiC introduces distinct generative artifacts. In particular, textures such as hair, grass, leaves, and mountains appear blurred, with competing methods failing to preserve sharp edges, leading to degraded visual quality. While PerCo produces visually realistic results, its ability to preserve fine-grained details (\textit{e.g.}, color, edges, shapes) is limited.
% undermining the goal of achieving high visual fidelity in image compression. 
In contrast, the proposed method, leveraging generative diffusion priors, not only produces sharp and realistic texture edges but also achieves higher visual fidelity in reconstructions. These results demonstrate its superior visual quality, validating its effectiveness in improving human-centered compression performance.
% In contrast, the proposed method, leveraging generative diffusion priors, not only provides sharper and more realistic texture edges, but also achieves high visual fidelity reconstructions. These results highlight its superior visual fidelity and subjective quality, validating its effectiveness in enhancing human-centered compression performance.

\subsection{Generalizability Evaluation}
The LDM used in this study is renowned for visual art generation, making the proposed generative coding method particularly suitable for AIGC artwork coding. To evaluate the method’s compatibility with various pretrained models, we apply another pretrained diffusion model, Realistic Vision V6.0, sourced from Civitai\footnote{https://civitai.com/models/4201/realistic-vision-v60-b1}, for compressing general photographic content. We finetune only the latent adapter $\mathcal{F}$ and the associated fusion module using the loss function in Eq.~(\ref{eq_chap5:loss_diffusion}) for 2 epochs, while keeping all other modules fixed.
% 
% To assess the proposed method's compatibility with different pretrained models, we apply
% the Realistic Vision V6.0 model for general photographed content compression, which is from Civitai\footnote{https://civitai.com/models/4201/realistic-vision-v60-b1} and trained on real-world data to generate realistic content (\textit{e.g.}, portraits, architecture, landscapes). We only finetune the latent adapter $\mathcal{F}$ and associated fusion module with Eq.~(\ref{eq_chap5:loss_diffusion}) for 2 epochs, with other modules all held fixed.
% The unified variable-rate encoder trained on the AIGC dataset, with only the latent adapter fine-tuned for 2 epochs.
% 
The proposed method is mainly compared against Cheng-VBR\cite{cheng2020learned} and TCM-VBR\cite{liu2023learned} on the Kodak dataset. Notably, the proposed method and Cheng-VBR shared the same encoder. 

As the RD results shown in Fig.~\ref{Fig:RD_all} (b) and Tab.~\ref{tab_chap5:BD-Rate}, the proposed method significantly outperforms HiFiC, PerCo, Cheng-VBR and TCM-VBR, achieving compression efficiency improvements of 81.6\% over Cheng-VBR and 69.9\% over TCM-VBR for in terms of BD-rate natural scene images. 
Additionally, our method achieves a reconstruction quality improvement of 45.8\% over Cheng-VBR and 29.3\% over TCM-VBR measured by BD-LPIPS.
% Our method achieves reconstruction quality improvement measured by BD-LPIPS by 45.8\% and 29.3\% over Cheng-VBR and TCM-VBR respectively.
% This demonstrates the method's versatility, compatibility with different pretrained diffusion models, and ability to generalize across various scenarios.
% 
For subjective quality at an 800$\times$ compression ratio shown in Fig.~\ref{Fig:sub_kodak}, the end-to-end learned coding methods exhibit significant blurring and distortion. In contrast, the proposed method maintains high visual fidelity, producing sharp, realistic textures. These results validate the effectiveness of the proposed generative coding approach in ensuring compatibility with various pretrained diffusion models and delivering strong performance across diverse scenarios.

\begin{table}[t]
% \vspace{-10pt}
\centering
\caption{Complexity analysis of the proposed method's modules and VVC reference software.}
\vspace{-3pt}
\label{tab:complexity}
\resizebox{1.0\linewidth}{!}{
% \begin{tabular}{c|c|c}
% \hline
% Method                & Module  & Inference Time (ms) \\ \hline
% \multirow{2}{*}{VVC}  & Encoder & 3940.6              \\ \cline{2-3} 
%                       & Decoder & 187.0               \\ \hline
% \multirow{3}{*}{Ours} & Encoder & 59.2                \\ \cline{2-3} 
%                       & Adapter & 2.6                 \\ \cline{2-3} 
%                       & LDM     & 76.5                \\ \hline
% \end{tabular}%
% }
\begin{tabular}{c|cc|ccc}
\hline
Method               & \multicolumn{2}{c|}{VVC}               & \multicolumn{3}{c}{Ours}                                          \\ \hline
Module              & \multicolumn{1}{c|}{Encoder} & Decoder & \multicolumn{1}{c|}{Encoder} & \multicolumn{1}{c|}{Adapter} & LDM  \\ \hline
Inference Time (ms) & \multicolumn{1}{c|}{3940.6}  & 187.0   & \multicolumn{1}{c|}{59.2}    & \multicolumn{1}{c|}{2.6}     & 76.5 \\ \hline
\end{tabular}}
\vspace{-14pt}
\end{table}

\subsection{Complexity Analysis}
Tab.~\ref{tab:complexity} presents the inference time for each module of the proposed method, with the LDM reflecting single-step inference. For reference, the encoding and decoding times of the VVC reference software VTM-11.0 on the CPU platform for $512\times 512$ resolution images are also provided. Notably, the adapter contributes minimally to the overall inference time. As a result, fine-tuning a lightweight adapter for different pretrained models introduces negligible computational overhead, offering an adaptable and computationally efficient solution for incorporating various generative diffusion priors. The reported compression performance of our method is based on 10 iterations of LDM inference. Despite increased inference time with more iterations, accelerated sampling research shows reduced complexity and fewer steps can yield quality images, enabling practical generative coding deployments.

% While multiple iterations increase inference time, ongoing research into accelerated sampling techniques for diffusion models suggests that high-quality images can be achieved with significantly reduced model complexity and fewer iterations, indicating promising potential for practical deployment of generative coding via diffusion priors.
% , even in a single step~\cite{zhao2023mobilediffusion}
\section{Conclusion}
This paper presents a generative coding framework that achieves high perceptual quality at low bitrates. By leveraging diffusion priors through a pre-optimized encoder, lightweight adapter, and fusion module, our method ensures compatibility with various pretrained diffusion models. The integration of attentive feature fusion and distribution renormalization further enhances reconstruction fidelity, improving compression efficiency. Experimental results demonstrate that the proposed method outperforms H.266/VVC by up to 79\%, showcasing its effectiveness and versatility across natural and AI-generated content. These findings highlight the method’s potential as an efficient solution for AI-generated content and a flexible approach to various generative coding applications.

\section*{Acknowledgment}
We thank Jie Wu, Hongbin Liu, Hao Yang and Siwei Ma for their insightful discussions and computational support.

\bibliographystyle{IEEEbib}
\bibliography{icme2025references}

% \vspace{12pt}
% \color{red}
% IEEE conference templates contain guidance text for composing and formatting conference papers. Please ensure that all template text is removed from your conference paper prior to submission to the conference. Failure to remove the template text from your paper may result in your paper not being published.

\end{document}